\documentclass[10pt,twocolumn,letterpaper]{article}

\usepackage[pagenumbers]{cvpr} 

\usepackage{graphicx}
\usepackage{amsmath}
\usepackage{amssymb}
\usepackage{booktabs}
\usepackage{bbm}
\usepackage{algorithm}
\usepackage{algpseudocode}
\usepackage{pifont}
\newcommand{\cmark}{\ding{51}}%
\newcommand{\xmark}{\ding{55}}%

\usepackage{multirow}
\usepackage{xcolor}

\usepackage{adjustbox}
\usepackage{array}
\usepackage{booktabs}
\usepackage{multirow}
\newcolumntype{R}[2]{%
    >{\adjustbox{angle=#1,lap=\width-(#2)}\bgroup}%
    l%
    <{\egroup}%
}
\newcommand*\rot{\multicolumn{1}{R{45}{1em}}}%

\usepackage[pagebackref,breaklinks,colorlinks]{hyperref}

\usepackage[capitalize]{cleveref}
\crefname{section}{Sec.}{Secs.}
\Crefname{section}{Section}{Sections}
\Crefname{table}{Table}{Tables}
\crefname{table}{Tab.}{Tabs.}

\def\cvprPaperID{7703} 
\def\confName{CVPR}
\def\confYear{2022}

\begin{document}

\title{MMPTRACK: Large-scale Densely Annotated Multi-camera Multiple People Tracking Benchmark}




\author{Xiaotian Han, \space
Quanzeng You, \space
Chunyu Wang, \space
Zhizheng Zhang, \space
Peng Chu, 
\\
Houdong Hu, \space
Jiang Wang, \space
Zicheng Liu
\\
Microsoft
\\
{\tt\small \{xiaothan, quyou, chnuwa, zhizzhang, pengchu, houhu, jiangwang, zliu\}@microsoft.com}
}

\maketitle

\begin{abstract}

Multi-camera tracking systems are gaining popularity in applications that demand high-quality tracking results, such as frictionless checkout because monocular multi-object tracking (MOT) systems often fail in cluttered and crowded environments due to occlusion. 
Multiple highly overlapped cameras can significantly alleviate the problem by recovering partial 3D information.
However, the cost of creating a high-quality multi-camera tracking dataset with diverse camera settings and backgrounds has limited the dataset scale in this domain.
In this paper, we provide a large-scale densely-labeled multi-camera tracking dataset in five different environments with the help of an auto-annotation system. 
The system uses overlapped and calibrated depth and RGB cameras to build a high-performance 3D tracker that automatically generates the 3D tracking results. The 3D tracking results are projected to each RGB camera view using camera parameters to create 2D tracking results. 
Then, we manually check and correct the 3D tracking results to ensure the label quality, which is much cheaper than fully manual annotation. 
We have conducted extensive experiments using two real-time multi-camera trackers and a person re-identification (ReID) model with different settings. This dataset provides a more reliable benchmark of multi-camera, multi-object tracking systems in cluttered and crowded environments. Also, our results demonstrate that adapting the trackers and ReID models on this dataset significantly improves their performance. 
Our dataset will be publicly released upon the acceptance of this work. 

\end{abstract}
\section{Introduction}
\label{sec:intro}

\begin{figure}[!htbp]
    \centering
    \includegraphics[width=0.475\textwidth]{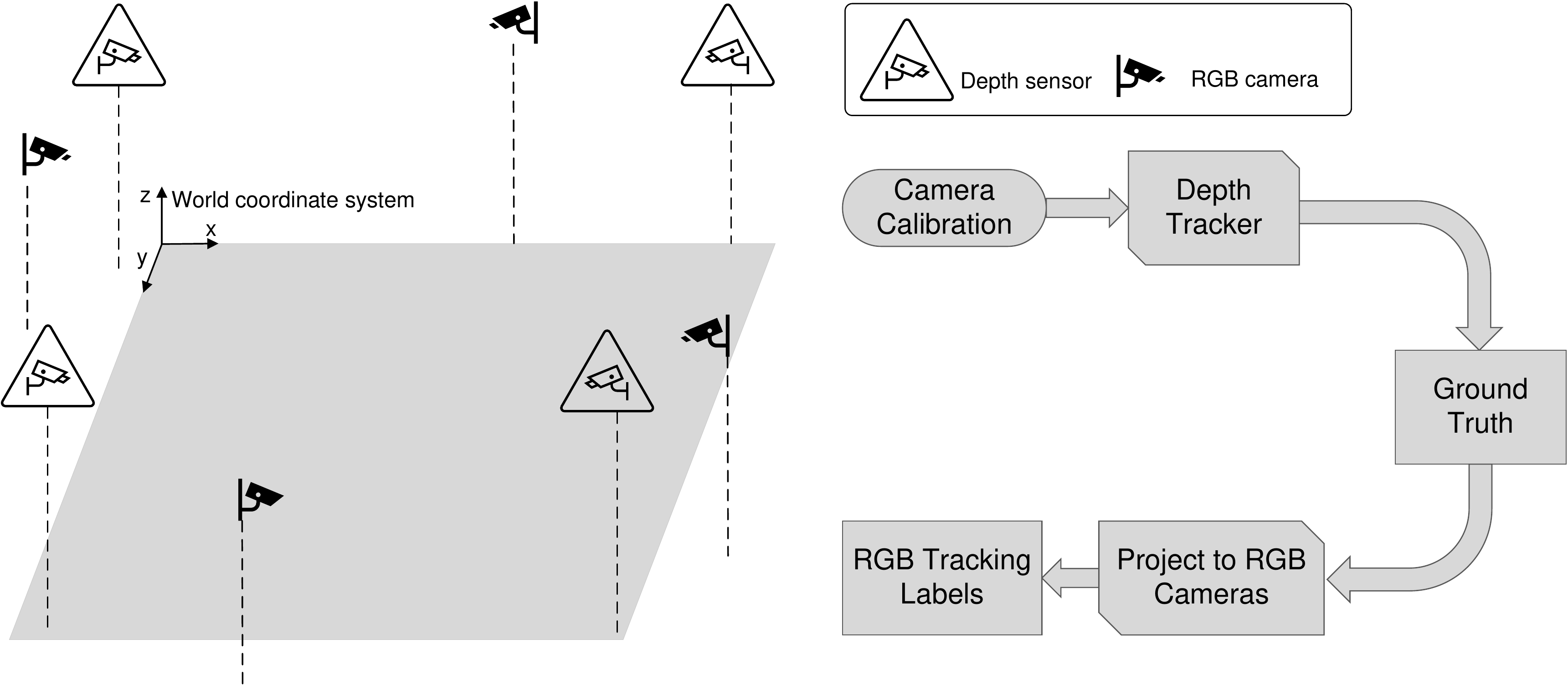}
    \caption{Illustration of our auto-annotation system. We have multiple calibrated depth sensors and RGB cameras installed in an environment. We build a high-performance 3D tracker to generate pseudo ground-truth tracking results in 3D space. These tracking results are projected to each RGB camera view as multi-camera multiple people tracking labels.}
    \label{fig:teaser}
    \vspace{-10pt}
\end{figure}

Multiple object tracking (MOT)~\cite{bernardin2008evaluating,luo2020multiple} is one of the fundamental research topics in computer vision. As more efficient and powerful deep neural networks are continuously being developed, the accuracy of MOT systems has been improved a lot in recent years.
However, monocular MOT systems still face severe challenges in cluttered and crowded environments, where occlusions of the tracked objects often occur. 
Thus, their accuracies are inadequate for applications that require highly accurate and consistent tracking results, such as frictionless checkout in retail stores or autonomous driving.

Recently, multi-camera systems have been widely deployed in these applications~\cite{amazongo}. 
The overlapped and calibrated cameras can considerably remedy the occlusion issue, and have achieved much higher accuracy than single-camera tracking systems~\cite{zhang2021voxeltrack}. 
However, we only have a few small multi-camera datasets publicly available due to data collection and annotation challenges. 
The lack of high-quality training and evaluation data makes it difficult to understand the challenges and improve the current multi-camera tracking systems.

In this paper, we propose a large-scale multi-camera multi-object tracking dataset, which is collected and annotated by an auto-annotation system. 
\figurename~\ref{fig:teaser} illustrates the overview of the auto-annotation system.
It consists of multiple calibrated depth sensors and RGB cameras. 
In particular, we follow the design in ~\cite{you2019action4d}, where the depth tracker works on the projected top-down view of the 3D space.
We train a top-down person detector on the projected view and follow the tracking-by-detection framework~\cite{andriluka2008people} to build a high-quality tracker on the projected view.
The 3D tracker can produce consistent and accurate tracking results on the projected top-down view.
We use human annotators to correct the 3D tracking errors, such as ID switches and false-positive tracks.
The corrected per-frame 3D tracking results are projected to all synchronized RGB streams using the camera parameters. 
Our experiments show that the auto-annotation system can produce very high quality tracking annotations (100\% IDF1 and 99.9\% MOTA) using only 1/800 of the cost of the traditional annotation methods. 



We set up five diverse and challenging environments equipped with the auto-annotation system in our lab. 
With the help of the auto-annotation system, we construct the largest multi-camera multiple people tracking dataset so far.
The dataset is densely annotated, \eg, per-frame bounding boxes and person identities are available. 

We evaluate two state-of-the-art real-time multi-camera person trackers~\cite{you2020real,zhang2021voxeltrack} and a person re-identification (ReID) model~\cite{zheng2016person} on this dataset under various settings.
Although the state-of-the-art multi-camera tracking systems perform much better than single-camera tracking systems, we find that their performance is still below the requirements of demanding applications.
In addition, the detectors, trackers, or Re-ID models trained on publicly available datasets, such as MS-COCO~\cite{lin2014microsoft} or MSMT~\cite{wei2018person}, are not performing well in these challenging environments because of large viewpoint and domain gap. Adapting the models using the training split of the data can significantly improve the accuracy of the system.
We expect the availability of such large-scale multi-camera multiple people tracking dataset will encourage more participants in this research topic.
This dataset is also valuable for the evaluation of other tasks, such as multi-view people detection~\cite{hou2020multiview,lima2021generalizable} and monocular multiple people tracking~\cite{bernardin2008evaluating}. 
To summarize, our contributions are as follows:
\begin{itemize}
    \item We construct the largest densely annotated multi-camera multiple people tracking dataset to encourage more research on this topic.
    \item We propose an auto-annotation system, which can produce high-quality tracking labels for multi-camera environments in a fast and cost-efficient way.
    \item We conduct extensive experiments to reveal the challenges and characteristics of our dataset.
\end{itemize}
\section{Related work}
\noindent \textbf{Approaches}\quad Multi-camera multi-object tracking has been extensively studied in the computer vision community. 
Previously, different graph-based approaches have been proposed to solve the data associations across different frames and  cameras~\cite{berclaz2011multiple,zamir2012gmcp,hofmann2013hypergraphs,tesfaye2017multi,wan2013distributed,chen2016equalized,wen2017multi,dehghan2015gmmcp,he2020multi}.
Recent approaches~\cite{ristani2018features,xu2019unified,vo2020self,hsu2020traffic} attempt to apply deep ReID features for the data association.
Extra efforts are needed to handle cross-camera appearance changes~\cite{jiang2018online,hou2019locality}. 
These methods can be applied to environments with non-overlapping cameras, but they cannot explicitly utilize the camera calibration information for cross-camera data association and 3D space localization.

Other approaches adopt camera calibration for tracklets merging and cross-camera association. Probabilistic occupancy map (POM)~\cite{fleuret2007multicamera} is one of the early representative studies. 
POM provides a robust estimation of the ground-plane occupancy, which is the key to building a high-performance tracker in crowded environments.
Also, homography~\cite{eshel2008homography} is employed to merge head segmentations from all camera views to build a head tracker. 
Later, deep occlusion~\cite{baque2017deep} extends this idea by utilizing Convolutional Neural Network (CNN) and Conditional Random Field (CRF) to reason the occlusions. 

Recently, 3D pose estimation and 3D person detection are also utilized for multi-camera person tracking~\cite{zhang2021voxeltrack}.
3D pose can be estimated~\cite{tu2020voxelpose,chen2020multi} by merging 2D skeleton estimations from multiple 2D camera views, using a 3D regression network or graph matching. 
Meanwhile, multi-view person detection approaches~\cite{hou2020multiview,lima2021generalizable,song2021stacked,hou2021multiview} also utilize camera calibration to merge multiple 2D detections or features to generate more reliable 3D person detection results. 
The accuracy of these approaches heavily depends on the quality of the 2D person detection or 2D pose estimation. 
These 3D poses and detections can be utilized for 3D trackers.

\begin{table}[!bp]
	\footnotesize
	\centering
	\vspace{-10pt}
	\begin{tabular}{@{\hskip 0.15mm}l@{\hskip 0.15mm}@{\hskip 0.15mm}c@{\hskip 0.15mm}@{\hskip 0.15mm}c@{\hskip 0.15mm}c@{\hskip 0.15mm}c@{\hskip 0.15mm}c|@{\hskip 0.15mm}c@{\hskip 0.15mm}c@{\hskip 0.15mm}}
		Dataset & \rot{\# of Envs} & \rot{Cameras} & FPS & \rot{Overlap} & \rot{Calib} & \shortstack{GT \\ (frames)} & \shortstack{Length \\ (minutes)}\\
		\toprule
		USC Campus\cite{kuo2010inter} & 1 & 3 &30 & No & No & 135,000 & 25 \\
		CamNet\cite{zhang2015camera} & 6 & 8 & 25 & Yes & No & 360,000 & 30 \\
		DukeMTMC\cite{ristani2016performance} & 1 & 8  & 60 & No & Yes & 2,448,000 & 85 \\
		SALSA\cite{DBLP:journals/corr/Alameda-PinedaS15} & 1 & 4 & 15 & Yes & Yes & $\sim$1,200 & 60 \\
		WILDTRACK\cite{chavdarova2018wildtrack} & 1 & 7 & 60 & Yes & Yes & $\sim$7$\times$9,518 & 60 \\
		\midrule
		MMPTRACK (ours) & 5 & 23 &15 & Yes & Yes & $\sim$2,979,900 & 576 \\
		\bottomrule
	\end{tabular}
	\caption{Representative multi-camera person tracking datasets. FPS stands for frame per second.}
	\label{table:datasets_comparison}
\end{table}
\noindent\textbf{Datasets}\quad Several multi-camera tracking datasets with highly overlapping cameras, have been widely adopted in multi-target multi-camera tracking research. 
Among them, PETS2009~\cite{ferryman2009pets2009}, Laboratory~\cite{fleuret2007multicamera}, Terrace~\cite{fleuret2007multicamera}, Passageway~\cite{fleuret2007multicamera}, USC Campus~\cite{kuo2010inter} and CamNet~\cite{zhang2015camera} have been 
collected with low-resolution cameras and these datasets only have a limited number of frames and person identities (IDs).
EPFL-RLC~\cite{chavdarova2017deep}, CAMPUS~\cite{xu2016multi} and SALSA~\cite{DBLP:journals/corr/Alameda-PinedaS15} are released more recently. However, EPFL-RLC only has 300 fully annotated frames, and CAMPUS comes without 3D ground truth. 
WILDTRACK dataset~\cite{chavdarova2018wildtrack} consists of high-quality annotations of both camera-view and 3D ground truth, as well as more person identities. 
However, the annotations are sparse and limited to 400 frames. 
DukeMTMC~\cite{ristani2016performance} is released with over 2 million frames and more than 2700 identities. However, there are almost no overlaps between different cameras. 
\tablename~\ref{table:datasets_comparison} compares our dataset (MMPTRACK) with several existing datasets. 
MMPTRACK is captured with a large number of calibrated overlapping cameras in indoor environments, which aligns better with the applications such as frictionless checkout. 
MMPTRACK is much larger than the existing data both in terms of the video length and the number of annotated frames.
The videos are densely labeled with per-frame tracking bounding boxes and person identities.


\section{Dataset collection}
This section gives a detailed overview of how we collect our dataset, including environment setup, camera calibration procedure, and annotation pipeline. 
\subsection{Dataset statistics}
\begin{table}[!htbp]
	\footnotesize
	\centering
	\begin{tabular}{l@{\hskip 0.5mm}|@{\hskip 0.75mm}c@{\hskip 0.75mm}|@{\hskip 0.75mm}c@{\hskip 0.75mm}|@{\hskip 0.75mm}c@{\hskip 0.75mm}|@{\hskip 0.75mm}c@{\hskip 0.75mm}|@{\hskip 0.75mm}c@{\hskip 0.75mm}|@{\hskip 0.75mm}c}
		\toprule
		\shortstack{Envs} & \shortstack{Retail} & \shortstack{Lobby} & \shortstack{Industry } & \shortstack{Cafe} & \shortstack{Office} & Total\\
		\midrule
		\# of cameras & 6 & 4 & 4 & 4 & 5 & 23\\
		\midrule
		Train (min) & 84 & 65 & 52 & 14 & 46 & 261 \\
		Validation (min) & 43 & 32 & 31 & 28 & 19 & 153 \\
		Test (min) & 45 & 32 & 32 & 31 & 22 & 162 \\
		\midrule
		Total (min) & 172 & 129 & 115 & 73 & 87 & 576 \\
		\bottomrule
	\end{tabular}
	\caption{Statistics of Multi-camera Multiple People Tracking (MMPTRACK) dataset.}
	\label{table:dataset_statistics}
\end{table}

The detailed statistics are summarized in \tablename~\ref{table:dataset_statistics}. 
Our dataset is recorded with 15 frames per second (FPS) in five diverse and challenging environment settings.
With the help of our auto-annotation system, we can obtain the per-frame tracking bounding box labels in the collected videos efficiently.
Overall, we collect about $9.6$ hours of videos, with over half a million frame-wise annotations for each camera view.
This is by far the largest publicly available multi-camera multiple people tracking (MMPTRACK) dataset.
\begin{figure}[!htbp]
	\centering
	\includegraphics[width=0.475\textwidth]{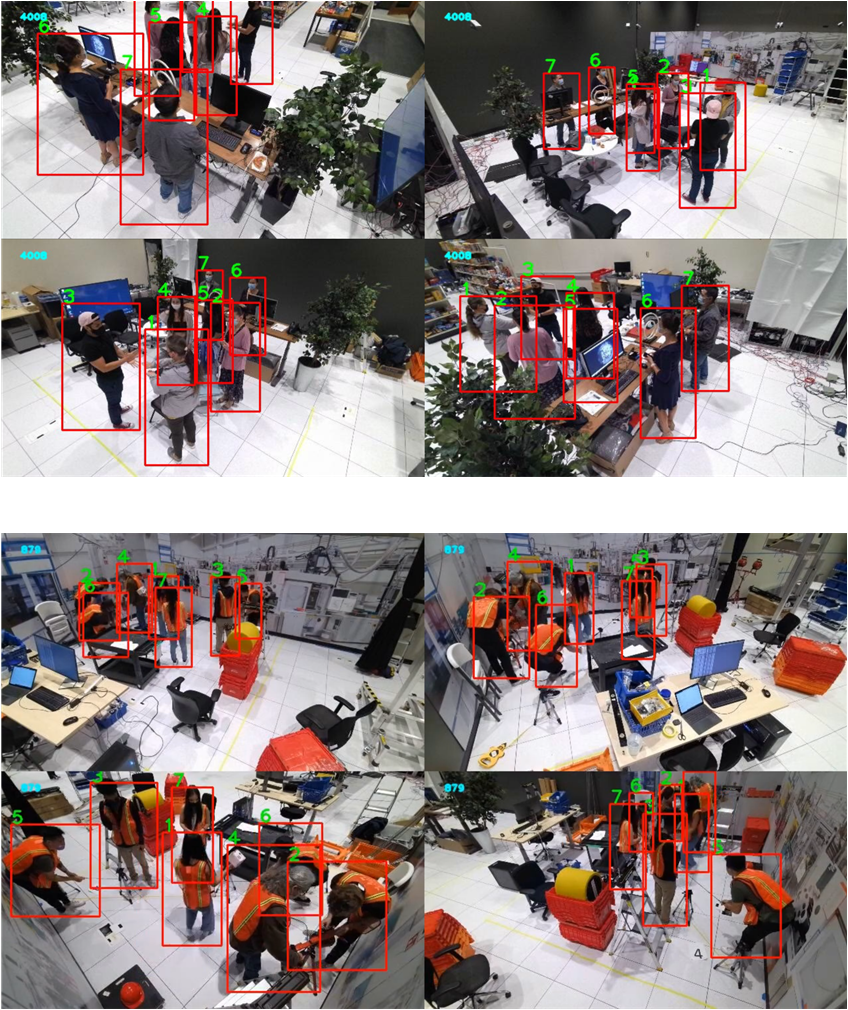}
	\caption{Examples of the images and tracking labels of our dataset. \textbf{Above}: images and labels of four camera views in \textit{Lobby} environment. \textbf{Below}: images and labels of four camera views in~\textit{Industry} environment.}
	\label{fig:dataset_visualization}
\end{figure}

\figurename~\ref{fig:dataset_visualization} shows some examples of the tracking labels of each camera view from two different environments. 
Although both environments are crowded and cluttered,
our ground truth consists of high-quality bounding boxes and consistent person IDs across all camera views.

\subsection{Environment setup}
\begin{figure}
	\centering
	\includegraphics[width=0.475\textwidth]{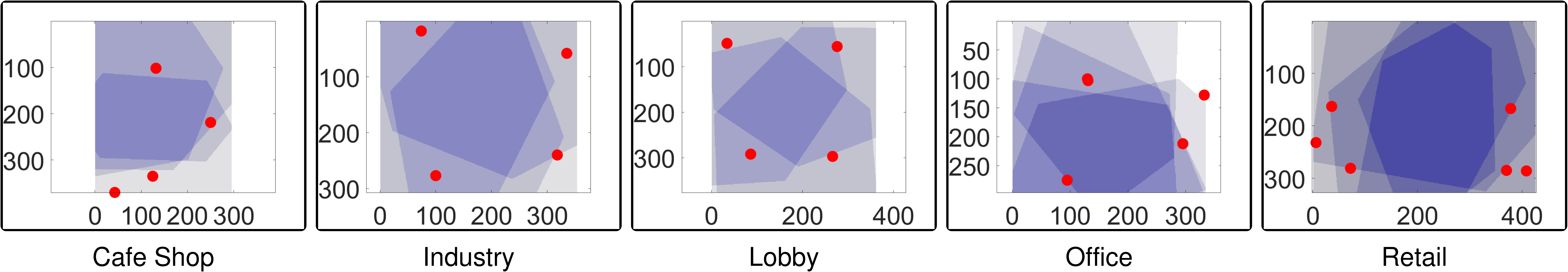}
	\caption{Field of view overlaps of different cameras in different environments on the ground plane. Each red dot represents the location of a camera. The X and Y axes represent the size of each environment in terms of pixels (each pixel unit is $20\mathit{mm}$).  
	}
	\label{fig:fov}
	\vspace{-10pt}
\end{figure}

We set up 5 different environments in our lab, \ie, \textit{Cafe Shop}, \textit{Industry}, \textit{Lobby}, \textit{Office} and \textit{Retail}.
We install Azure Kinect cameras in each of these environments with fixed positions and camera angles. 
\figurename~\ref{fig:fov} shows the field of views overlaps among different cameras on the ground plane. 
Azure Kinects can record RGB and depth streams simultaneously. 
Azure Kinects' RGB streams will be used as the default RGB cameras for our dataset (see \figurename~\ref{fig:teaser})\footnote{With multiple camera calibrations, 3D tracking labels can be projected to other types of RGB cameras.}.
The depth and RGB streams of all Azure Kinects are all recorded and time-synchronized.


\subsection{Camera calibration}

\begin{figure}[!htbp]
	\centering
	\includegraphics[width=0.4\textwidth]{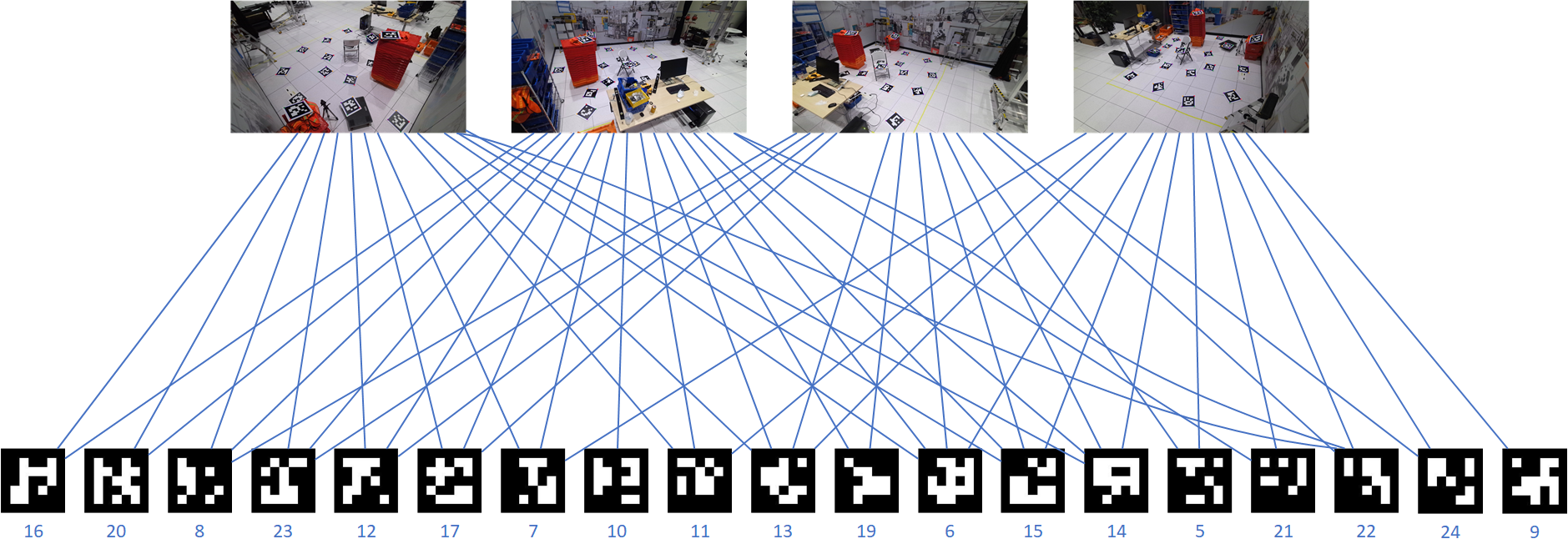}
	\caption{Example of bipartite graph from cameras and ArUco markers. The above 4 cameras are from Industry environment. }
\label{fig:camera_graph}
\vspace{-10pt}
\end{figure}

\noindent\textbf{Intrinsic parameters} We obtain Azure Kinect intrinsic parameters directly from its SDK. We denote intrinsic parameters as \(I\).

\noindent\textbf{Extrinsic parameters} In our settings, one camera has the overlapping field of views with at least another camera. 
We use ArUco markers as reference points in the world coordinate system. 
We build a connected bipartite graph, where cameras and ArUco markers are vertices. 
If ArUco marker $m_i$ is within the view of camera $c_j$, we will add an edge $e_{ij}$ between them.
\figurename~\ref{fig:camera_graph} shows an example of the connected bipartite graph when we calibrate the \textit{Industry} environment.
Let \(P=\bigcup{P_i}\) be the set of detected corner points of all markers ($P_i$ is the corners from $i$-th marker).
Then, the set of extrinsic parameters $E$ is obtained by optimizing
\begin{equation} 
\label{eq:1}
E^* = \arg\max_{E, M} \sum_{i=1}^{|P|} \sum_{c=1}^{C} \mathbbm{1} _i ^c \|p_i^c - I^c*E^c*m_i\|^2,
\end{equation}
where \(\|\cdot\|\) denotes Euclidean distance, 
\(\mathbbm{1} _i ^c\) is an indicator function, whose value equals to 1 only if point \(p_i\) is visible in camera view \(c\), \(M =\{m_i, i = 1,\cdots,|P|\} \)
denotes the markers' corner points in world coordinate system. 
The graph optimization approach proposed in~\cite{5979949} is implemented to solve Eq.~(\ref{eq:1}).

\subsection{Dataset collection}
The current dataset is recorded in four half-day sessions.
In each session, we hire seven different subjects to participate. 
The actors can act improvisationally as long as their action fits the environment setting. 
For instance, in \textit{Retail} environment, they are free to perform any shopping behaviors, \eg, pushing shopping carts, holding baskets, and standing in a queue for checkout. 
Following such instructions, we believe the collected dataset covers a wide variety of behaviors. 
In total, we have 28 subjects, with different ages, genders, and ethics, who participated in the data collection (see \tablename~\ref{table:dataset_statistics} for the statistics of our dataset).

\subsection{3D auto-annotation system}

Our 3D auto-annotation system utilizes depth streams to track multiple people with high accuracy.
The overall workflow of the system is described in Algorithm~\ref{alg:autoannotation_workflow}.
We build our 3D tracker using data released in~\cite{you2019action4d}, which does not have any overlap with the current dataset in terms of environment or subject. 
The top-down view image is constructed from the merged 3D point cloud. 
This design avoids environment-dependent factors, such as lighting, camera angles, \etc.
Therefore, the 3D tracker can be easily generalized and applied to different environments.


\begin{algorithm}
\footnotesize
\caption{Workflow of RGBD Auto-annotation System}
\label{alg:autoannotation_workflow}
\textbf{Input:} Synchronized RGB and depth steams and camera parameters $C$\\
\textbf{Output:} Person bounding boxes and IDs in each camera view 
\begin{algorithmic}
\Procedure{Auto-Annotation}{}
\State $B\leftarrow list()$ \Comment{Person bounding boxes}
\While{All Streams not end}
\State $R \leftarrow set()$ \Comment{Synchronized RGB images}
\State $D \leftarrow set()$ \Comment{Synchronized Depth images}
\For{\texttt{stream in Streams}}
\State $r, d \leftarrow \text{stream.read}()$
\State \textbf{add}($r$, $R$)
\State \textbf{add}($d$, $D$)
\EndFor
\State $P \leftarrow$ \textbf{PointCloudGen}($R$, $D$, $C$)
\State $T_d \leftarrow$ \textbf{TopdownViewGen}($P$) \Comment{Top-down view of the scene}
\State $B \leftarrow$ \textbf{PersonDetector}($T_d$)
\State $T_r \leftarrow$ \textbf{3DTracker}($B$, $P$) \Comment{3D Tracklets}
\State $B_c \leftarrow$ \textbf{Projection}($T_r$, $C$) \Comment{Camera-view bounding boxes}
\State \textbf{append}($B_c$, $B$)
\EndWhile
\State \Return $B$
\EndProcedure
\end{algorithmic}
\end{algorithm}

\noindent\textbf{Point cloud reconstruction}\quad We reconstruct the point cloud of the whole scene from calibrated and synchronized depth cameras.
Given the intrinsic parameters $I$ and extrinsic parameters $E$, the point cloud $\mathbb{P}$ is calculated as follows:
\begin{equation} 
\label{eq:2}
\mathbb{P} = \bigcup_{c=1}^{C} \bigcup_{i} \bigcup_{j} (E^c)^{-1} * (I^c)^{-1} * \left[i, j, d^c_{i,j}\right]^T 
\end{equation}
where $i$ and $j$ index over all valid locations and \(d_{ij}^c\) denotes camera \(c\)'s depth measurement at location \((i, j)\).

\noindent\textbf{Top-down view projection}\quad
\label{sec:td:proj}
We discretize the point cloud $\mathbb{P}$ into a binary voxel set $\mathbb{V}$.
Each voxel $\mathbb{V}_{i,j,k}$ covers a cube with a volume of $20 \mathit{mm} \times 20 \mathit{mm} \times 20\mathit{mm}$.
$\mathbb{V}_{i,j,k} = 1$ if and only if there exists at least one point $\mathbb{P}_{i',j',k'}$, such that it locates inside the cube covered by $\mathbb{V}_{i,j,k}$.

We set the world-coordinate system's X and Y axes parallel to the ground plane and the Z axis vertical to the ground. 
The top-down view image $T_d$ can be obtained by projecting $\mathbb{V}$ onto the X-Y ground plane.
More specifically, its value at position \((m, n)\) is computed as:
\begin{equation} 
\label{eq:3}
T_d(m, n) = \underset{z, V(m,n,z) = 1}{\arg\max} \mathbb{V}_{m,n,z},
\end{equation}
which can be perceived as the \textit{height} of filled voxels within each cube $V_{m,n,(\cdot})$.

\noindent\textbf{Top-down view person detection}\quad
We design a simple two-stage top-down person detector.
We assume that each person center is roughly the highest point around a local region. 
Recall that pixel values of $T_d$ (Eq.~(\ref{eq:3})) represent the height of each location. 
Therefore, in the proposal generation stage, we extract all local maxima from the top-down view image $T_d$. 
For each candidate position $(i,j)$, we crop a \(20 \times 20\) square region centered around it.  
The cropped image region is fed into a Convolutional Neural Network, which serves as a person classifier in the second stage.

\noindent\textbf{3D tracker}\quad We develop a highly efficient 3D association method. 
The inputs of our method are top-down view detection boxes with corresponding detection scores and point clouds. 
We initialize a tracklet with a detection box whose score is above a threshold at the beginning stage.
Then, for the following frames, 
we construct the cost matrix 
based on spatial and appearance (color histogram) distance between each tracklet and the detected bounding boxes. 
Association results are obtained by employing Hungarian Matching algorithm. 
For each unmatched detection bounding box, we generate a new candidate tracklet.

\noindent\textbf{Camera-view projection}\quad
The height $h$ of each tracked person can be estimated from the local maxima of its top-down bounding box (each pixel value unit represents $20\mathit{mm}$).  
We fit a cube with a size of $100\mathit{cm}\times100\mathit{cm}\times h$ bottom centered at each top-down tracked person.
A 3D bounding box (cube) is projected to each camera view. 
The 2D bounding box in each view is the tightest rectangle that encloses the projected 3D bounding box in this view.
In this way, we propagate the tracking results from 3D space to all RGB cameras within the same environment.

\subsection{Annotation and quality control} 

The 3D tracker may still introduce errors occasionally. 
We manually fix all tracking errors before propagating the results to each RGB camera view.
The most common errors in our RGBD Tracker are tracklet ID switch and false-positive person detection. 
We request annotators to correct ID switches and remove false-positive tracklets from 3D tracking results. 
Notice this process is relatively cost-efficient because no bounding box labeling is required, and all the corrections are performed at the tracklet level.


Based on our experience, each annotator can label around 600 frames (including boxes and IDs) per day for videos with around 5 to 6 persons inside. 
Manually labeling all the videos in our dataset costs 414 labeler days when we annotate every ten frames and interpolate the tracking labels to the remaining frames. 
Labeling all the frames costs more than 4000 labeler days. 
In comparison, we only need one labeler to work less than 5 hours to correct all the errors of our 3D tracker. 

\begin{table}[!tbp]
\footnotesize
\centering
\begin{tabular}{l|c|c|*{5}c}
\toprule
Envs &  IDF1$\uparrow$ & MOTA$\uparrow$ & FP$\downarrow$ & FN$\downarrow$ & IDs$\downarrow$\\
\midrule
Cafe  & 100 & 100 & 0 & 0 & 0 \\
Industry  & 100 & 100 & 0 & 0 & 0 \\
Lobby & 100 & 100 & 0 & 0 & 0 \\
Office & 100 & 100 & 0 & 0 & 0\\
Retail & 100 & 99.9 & 0 & 4 & 0 \\
\bottomrule
\end{tabular}
\caption{Performance of our 3D tracker on one testing sequence.}
\label{table:perf:3dtracker}
\vspace{-10pt}
\end{table}
To test the quality of the 3D tracking ground truth, we sample $1,000$ continuous frames from each environment and manually label all the bounding boxes and ids of each tracklet. 
\tablename~\ref{table:perf:3dtracker} summarizes the evaluation results of our corrected 3D tracker's results using human-labeled ground truth.
Only four human-labeled targets are not matched by our 3D tracking results, which is tolerable given that humans can also make errors.

\section{Benchmarks}
In this section, we discuss the evaluation metrics, evaluated approaches and experimental results on both \textit{tracking} and \textit{ReID} tasks.
\subsection{Evaluation metrics} 
For the tracking task, we follow the widely adopted MOT metrics to evaluate the performance~\cite{bernardin2008evaluating}. 
We report the false positive (\textbf{FP}) and false negative (\textbf{FN}) detections, which are also considered in multiple object tracking accuracy (\textbf{MOTA}). 
MOTA further deals with identity switches (\textbf{IDs}) and is extensively used in benchmarking different trackers.
Besides, we also report \textbf{IDF1}, which measures the ID consistency between the predicted trajectories and the ground truth using ID precision and ID recall.
IDF1 can assess the trackers' ability on tracklet association. 
We report all performance metrics on the top-down view for multi-camera tracking, using the 3D ground-truth tracking results. 
We follow the settings in~\cite{chavdarova2018wildtrack}, where a radius of one meter is used as the distance threshold when computing all metrics.

For the ReID task, we adopt the widely used Rank-1 accuracy (\textbf{R-1}) and mean Average Precision (\textbf{mAP})~\cite{zheng2016person} to compare the model's performance under different settings.
\subsection{Evaluation approaches}
\subsubsection{Baselines trackers}
We evaluate two state-of-the-art online real-time multi-camera trackers on our datasets. 
We focus on evaluating online real-time trackers because they can better reflect the core detection and tracking performance, and
we can better observe the challenges in our dataset in these evaluations.

\noindent\textbf{End-to-end deep multi-camera tracker (DMCT)} \quad 
In this baseline, we employ the end-to-end approach (DMCT) proposed in~\cite{you2020real}. 
This approach estimates the ground point heatmap of each candidate at each camera view, projects 
the ground point heatmaps from all camera views to the ground plane, and fuses all the heatmaps into a ground-plane heatmap.
Similar to~\cite{you2020real}, we train a variant of CornerNet~\cite{law2018cornernet} with pixel-wise Focal Loss~\cite{law2018cornernet} as our ground-point estimation model.
The tracker works on the fused ground-plane heatmap. 

\begin{figure}[!htbp]
	\centering
	\includegraphics[width=0.4\textwidth]{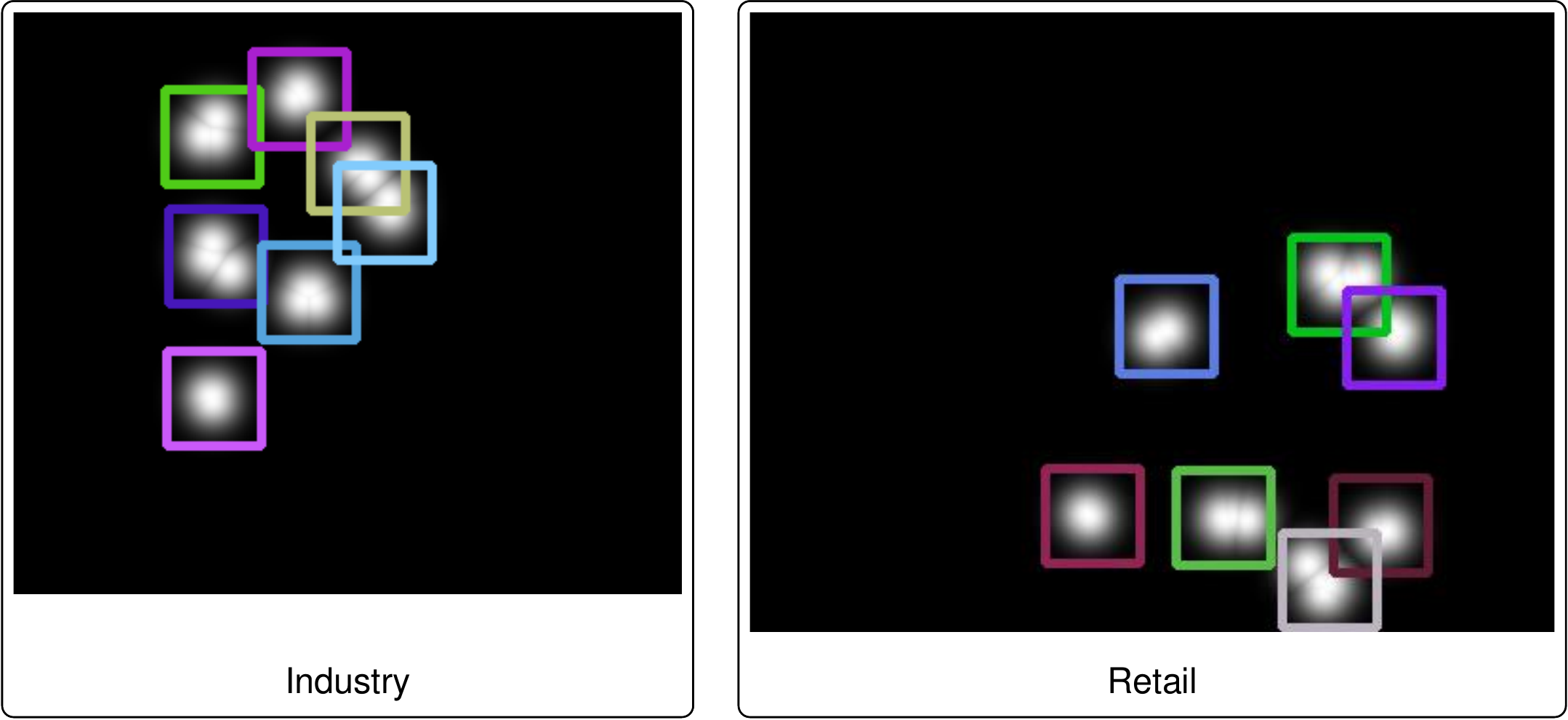}
	\caption{Examples of the fused ground-plane images as inputs to person detector. The bounding boxes are the ground truth.}
	\label{fig:td}
	\vspace{-15pt}
\end{figure}

Given the fused ground-plane heatmap $H$, two different approaches are utilized to obtain top-down person detections. 
The first approach is rule-based. It directly applies Gaussian blur to $H$ and extracts local maxima as person detections for tracking. 
In this approach, The heatmaps $\{H_1, H_2, \cdots, H_C\}$ from $C$ camera views are projected to the ground plane using homographies between the ground plane and all camera views. For each location in fused top-down heatmap $H$, its value is the maximum over all camera view's projected heatmap.

The second approach trains a YOLOV5~\cite{glenn_jocher_2021_5563715} detector as top-down person detector. 
The second approach is more expensive than the first approach, but it is much cheaper than the sequence-based deep glimpse network in~\cite{you2020real}.
In this approach, we first find the local maxima at each camera-view heatmap $H_i$ as candidate points.
These points are projected to the ground plane, and each point generates a Gaussian distribution around it to reduce noise. 
We keep the maximum value from all Gaussian distributions for each location in the projected top-down image. 
\figurename~\ref{fig:td} shows two examples of fused ground-plane heatmaps from two different environments. 
Although the two environments are configured with different RGB camera settings and backgrounds, the top-down images look similar.
Our experimental results will show that the top-down detector can easily generalize across different environments.  


We also label another external dataset using images from OpenImage\footnote{\url{https://storage.googleapis.com/openimages/web/index.html}}. 
Specifically, we sample a subset of OpenImage containing persons, which has about $600,000$ images.
We manually label the ground point of each person in these images. 
In our experiments, we evaluate the impact of this external data on tracking performance.

The variants of DMCT approach include: \textbf{DMCT} train the ground-point estimation model on the training split of MMPTRACK and the rule-based approach for top-down person detection; \textbf{DMCT-TD} uses the same ground-point estimation model with DMCT and the deep learning-based top-down person detector; \textbf{DMCT-Ext} uses the same rule-based top-down person detection with DMCT, and it trains ground-point estimation model with both training split of MMPTRACK and the extra manually labeled OpenImage dataset; \textbf{DMCT-Ext-TD} uses the same ground-point estimation model with DMCT-Ext, and the deep learning-based top-down person detector.

\noindent \textbf{Tracking by 3D skeletons (VoxelTrack)} \quad
This baseline performs tracking with estimated 3D body joints, which have more spatial information than the single ground point. 
It is built on top of a state-of-the-art 3D pose estimation method VoxelPose \cite{tu2020voxelpose}. 
It requires neither camera-view $2$D pose estimation nor cross-camera pose association as in previous works, which is error-prone. 
Instead, all hard decisions are postponed and made in the $3$D space after fusing $2$D visual features from all views, which effectively avoids error accumulation. 
In addition, the fused representation is robust to occlusion. A joint occluded in one camera view may be visible in other camera views. 

We follow a standard pipeline \cite{zhang2021fairmot} for tracking the 3D poses. 
We initialize every estimated $3$D pose as a tracklet. 
For the following frames, we use the Hungarian algorithm to assign the $3$D poses to the existing tracklets, 
where the matching cost is the sum of the 3D distance for all the 3D joints.
We reject the assignment if the spatial distance between the tracklet and the $3$D pose is too large. 
An unmatched $3$D pose will be assigned to a new tracklet. 
An existing tracklet will be removed if it is not matched to any $3$D poses for more than $30$ frames. 

Following the settings of \cite{tu2020voxelpose}, the 2D heatmap estimation model is trained on the COCO dataset, and the 3D models are trained using synthetic data. 
Since MMPTRACK lacks $3$D pose labels, 
we conduct a \textit{virtual} fine-tuning for the 3D models.  
Calibration parameters of MMPTRACK are employed to generate pseudo-3D human poses. 
\vspace{-10pt}
\subsubsection{Baseline ReID models}
\label{sec:baseline:reid}
We evaluate an advanced person re-identification (ReID) model proposed in FastReID~\cite{he2020fastreid} on the MMPTRACK dataset. 
The robust ReID feature can improve the performance of an MOT system. 
We study the challenges of learning discriminative ReID features in a cluttered and crowded environment under multiple cameras.
Our baseline model is built upon a commonly used baseline model \cite{luo2019bag}. 
We further incorporate Non-local block \cite{wang2018non}, GeM pooling \cite{berman2019multigrain} and a series of training strategies (see details in \cite{he2020fastreid}). 

We uniformly downsample the MMPTRACK dataset with a step size of 32. 
During testing, we divide each down-sampled sequence into two halves. 
We use the cropped persons in the first half as the query set and those in the second half as the gallery set. 
Although there are only a small number of person identities in this dataset, the diverse camera angles and cluttered backgrounds still make ReID a challenging task on our dataset. 

We evaluate three training configurations of the above model on the testing split of MMPTRACK. 
Specifically, for the \textbf{Generalization} setting, we directly evaluate the model trained with the person ReID dataset MSMT~\cite{wei2018person}. 
For the \textbf{Adaptation} setting, we perform supervised fine-tuning of the previous model with cropped persons on the training split of MMPTRACK. 
For the \textbf{Supervised} setting, we train the person ReID model from scratch using only the cropped persons in the training split of MMPTRACK.

\subsection{Benchmark results and discussions}
\subsubsection{Tracking performance on MMPTRACK}
We evaluate the two real-time baseline trackers on the collected MMPTRACK dataset.


\begin{table}[!htbp]
	\scriptsize
	\centering
	\begin{tabular}{l|*{5}c}
		\toprule
		Method  & IDF1$\uparrow$ & MOTA$\uparrow$ & FP$\downarrow$ & FN$\downarrow$ & IDs$\downarrow$\\
		\midrule 
		VoxelTrack & 55.2 & 79.6 & 43,776 & 110,239 & 4,365 \\
		\midrule
		DMCT & 60.2 & 91.5 & 34,450 & 41,920 & 2,158 \\
		DMCT-TD & 74.8  & 93.6 & 15,080& 42,854 & 620 \\
		DMCT-Ext  & 61.1 & 92.5 & 30,789 & 36,631 & 1,953 \\
		DMCT-Ext-TD & 77.5 & 94.8 & 19,235 & 28,505 & 567 \\
		\bottomrule
	\end{tabular}
	\caption{Tracking performance on validation split.}
	\label{table:tracking:val}
	\vspace{-10pt}
\end{table}

\begin{table}[!htbp]
	\scriptsize
	\centering
	\begin{tabular}{l|*{5}c}
		\toprule
		Method  & IDF1$\uparrow$ & MOTA$\uparrow$ & FP$\downarrow$ & FN$\downarrow$ & IDs$\downarrow$\\
		\midrule 
		VoxelTrack & 50.8 & 76.8&  49,881 & 142,380& 4,922 \\
		\midrule
		DMCT & 56.0 & 88.8 & 39,715 & 52,559 & 2,677 \\
		DMCT-TD & 68.1  & 93.2 & 16,023 & 40,606 & 935 \\
		DMCT-Ext  & 56.6 & 89.0 & 42,413 & 48,039 & 3,013 \\
		DMCT-Ext-TD & 74.1 & 94.6 & 7,005 & 38,296 & 641 \\
		\bottomrule
	\end{tabular}
	\caption{Tracking performance on testing split.}
	\label{table:tracking:test}
	\vspace{-10pt}
\end{table}


\begin{table*}[!htbp]
	\footnotesize
	\centering
	\begin{tabular}{l|c|c|*{5}{c}|*{5}{c}}
		\toprule
		\multirow{2}{*}{Method} & \multirow{2}{*}{w/Env Training} &\multirow{2}{*}{Env} & \multicolumn{5}{c|}{Without external data} & \multicolumn{5}{c}{With external data}\\
		\cline{4-13}
		& & & IDF1$\uparrow$ & MOTA$\uparrow$ & FP$\downarrow$ & FN$\downarrow$ & IDs$\downarrow$ & IDF1$\uparrow$ & MOTA$\uparrow$ & FP$\downarrow$ & FN$\downarrow$ & IDs$\downarrow$\\
		\midrule
		DMCT & \xmark & Cafe & 39.4 & 87.7 & 12,012 & 7,589 & 1,158 & 56.9 & 91.8 & 6,792 & 6,382 & 701 \\
		DMCT & \cmark & Cafe & 64.2 & 95.9 & 2,691 & 4,063 & 162  & 61.3 & 96.0 & 2,104 & 4,409 & 297\\
		\midrule
		DMCT & \xmark & Industry  & 34.2& 78.4 & 18,486 & 20,548 & 1,637  & 42.7 & 82.7 & 11,700 & 19,858 & 962  \\
		DMCT & \cmark & Industry  & 61.7& 90.5 & 9,107 & 8,431 & 306 & 64.5& 91.2 & 9,188 & 7,155 & 233\\
		\midrule
		DMCT & \xmark & Lobby & 47.1 & 86.4 & 13,774 & 12,223 & 1,136 & 50.6 & 89.6 & 12,574 & 7,425 & 720 \\
		DMCT & \cmark & Lobby & 69.4 & 94.5 & 3,343 & 7,361 & 318 & 69.2 & 95.1 & 2,445 & 7,007 & 247\\
		\midrule
		DMCT & \xmark & Office  & 50.0 & 89.0 & 3,287 & 8,577 & 591 & 40.0 & 85.8 & 2,849 & 12,569 & 735 \\
		DMCT & \cmark & Office  & 68.0 & 93.7 & 1,514 & 5,625 & 67 &66.8 & 93.9 & 1,454 & 5,304 & 127\\
		\midrule
		DMCT & \xmark & Retail & 27.7 & 60.7 & 79,443 & 15,788 & 3,170 & 30.4 & 70.7 & 50,713 & 20,060 & 2,667 \\
		DMCT & \cmark & Retail & 45.7 & 85.8 & 17,795 & 16,440 & 1,305 & 49.4 & 88.3 & 15,598 & 12,756 & 1,049 \\
		\bottomrule
	\end{tabular}
	\caption{Tracking performance of each environment on validation split. We compare the detection model trained with and without each domain-specific data. Without any environment-specific data, trackers' performance drops significantly. }
	\label{table:tracking:mmp:ablation}
\end{table*}

\tablename~\ref{table:tracking:val} and \tablename~\ref{table:tracking:test} include the results of different baseline trackers on the validation and testing splits of our dataset respectively.
The results suggest that the top-down person detector trained with a deep learning model can significantly boost the performance of baseline DMCT, especially for \textbf{IDF1} and \textbf{IDs}.
With an extra of $600,000$ images from OpenImage, \textbf{DMCT-Ext} shows slight improvements over \textbf{DMCT} in terms of \textbf{IDF1} and \textbf{MOTA}. 
\textbf{DMCT-Ext-TD} improves \textbf{IDF1} by  $2.7$ over \textbf{DMCT-TD}.
However, the \textbf{MOTA} only increases $1.1$.
Compared with \textbf{VoxelTrack}, which is only virtually fine-tuned on MMPTRACK, all variants of \textbf{DMCT} perform better. 
We believe the performance gap is due to the large domain differences of our dataset and other public datasets such as MS-COCO, on which the \textbf{VoxelTrack} is trained. 
Since we can easily generate a large-scale multi-camera multi-object tracking dataset for an environment using our auto-annotation system, we can train a model that adapts to a given environment with improved accuracy. 
However, the accuracy of the baseline methods is still not high enough to meet the requirements of demanding applications, particularly for IDF1. Further research is needed in this domain.


\noindent \textbf{Ablation studies} \quad 
We study the impact of different training splits on our baselines' performance. 
Since \textbf{VoxelTrack} can only be virtually fine-tuned on our dataset, we only cover the ablation study results of different variants of \textbf{DMCT}. 
We attempt to study the impact of environment-specific data. 
In particular, we train the ground point estimation models and the top-down detectors with and without each environment-specific data.
Then, we report the performance in each environment individually.

\tablename~\ref{table:tracking:mmp:ablation} shows the tracking evaluation metrics without the top-down detector.
Generally, without each environment's data, the tracking performance drops significantly.
This is particularly true in terms of the IDF1 metric. 
Also, external training data can improve the tracker's performance for most environments when environment-specific data is absent.
However, with environment-specific data, external data leads to a limited performance gain. 
\begin{table*}[!htbp]
	\footnotesize
	\centering
	\begin{tabular}{l|c|c|*{5}{c}|*{5}{c}}
		\toprule
		\multirow{2}{*}{Method} & \multirow{2}{*}{w/Env Training} &\multirow{2}{*}{Env} & \multicolumn{5}{c|}{Without external data} & \multicolumn{5}{c}{With external data}\\
		\cline{4-13}
		& & & IDF1$\uparrow$ & MOTA$\uparrow$ & FP$\downarrow$ & FN$\downarrow$ & IDs$\downarrow$ & IDF1$\uparrow$ & MOTA$\uparrow$ & FP$\downarrow$ & FN$\downarrow$ & IDs$\downarrow$\\
		\midrule
		DMCT-TD & \xmark & Cafe & 77.4 & 95.7 & 503 & 6,635 & 39 & 76.0 & 96.8 & 488 & 4,885 & 36  \\
		DMCT-TD & \cmark & Cafe & 76.4 & 96.9 & 740 & 4,385 & 62 & 74.8 & 97.1 & 742 & 4,119 & 53 \\
		\midrule
		DMCT-TD & \xmark & Industry  & 73.8 & 87.7 & 7,400 & 15,661 & 62 & 74.2 & 90.0 & 7,714 & 11,039 & 67 \\
		DMCT-TD & \cmark & Industry  & 79.0 & 91.1 & 7,692 & 8,947 & 64 & 79.4 & 92.6 & 8,021 & 5,812 & 47 \\
		\midrule
		DMCT-TD & \xmark & Lobby & 88.4 & 96.4 & 520 & 6,587 & 83 & 82.4 & 97.2 & 115 & 5,419 & 54 \\
		DMCT-TD & \cmark & Lobby & 85.7 & 96.2 & 31 & 7,520 & 49 & 86.8 & 97.3 & 145 & 5,303 & 25 \\
		\midrule
		DMCT-TD & \xmark & Office  & 85.2 & 93.7 & 714 & 6,420 & 43 & 85.9 & 97.6 & 884 & 1770 & 38 \\
		DMCT-TD & \cmark & Office  & 81.3 & 97.4 & 787 & 2,182 & 42 &  89.0 & 98.0 & 994 & 1,237 & 47 \\
		\midrule
		DMCT-TD & \xmark & Retail & 56.4 & 87.7 & 8,622 & 21,549 & 592 & 57.8 & 85.9 & 13258 & 21397 & 747  \\
		DMCT-TD & \cmark & Retail & 58.5 & 89.6 & 5,820 & 19,820 & 403 & 65.3 & 91.3 & 9333 & 12034 & 395 \\
		\bottomrule
	\end{tabular}
	\caption{Tracking performance of each environment on validation split with the top-down detector. We compare the detection model trained with and without each domain-specific data. With the top-down detector, the performance of different models is similar.}
	\label{table:tracking_td:mmp-oi:ablation}
	\vspace{-10pt}
\end{table*}

\tablename~\ref{table:tracking_td:mmp-oi:ablation} further studies DMCT's performance when equipped with a deep learning-based top-down detector.
A tracker with a deep learning-based top-down detector has better generability. 
Without external data, models trained with environment-specific data show better performance in \textit{Industry} and \textit{Retail} in terms of \textbf{IDF1}. 
However, models trained without environment-specific data even show better \textbf{IDF1} in \textit{Cafe}, \textit{Lobby} and \textit{Office}.
Also, when extra OpenImage data is utilized to train a ground point estimation model, the performance gain is limited, and in some environments, even worse than the results without the external data. 
It is generally believed that a pre-trained model on external data may provide good initialization when training the deep model. 
However, the domain gap between MMPTRACK and OpenImage makes the pre-training step insignificant. 
Instead, the large-scale in-domain MMPTRACK dataset can train a model with better performance. 
Meanwhile, compared with \tablename~\ref{table:tracking:mmp:ablation}, the results in \tablename~\ref{table:tracking_td:mmp-oi:ablation} also suggests that the deep-learning-based top-down detector reduces the performance gap caused by external ground-point data. 

\subsubsection{ReID performance on MMPTRACK}
We report the results of the ReID model with the three different settings discussed in Section~\ref{sec:baseline:reid}. 
The MSMT ReID dataset, which is employed to pretrain our \textbf{Generalization} and \textbf{Adaptation} model, consists of more than $4,000$ indoor and outdoor person identities. 
The evaluation results are summarized in \tablename~\ref{tab:reid}. 
Even though our training dataset consists of only $14$ different person identities, training from scratch still outperforms the \textbf{Generalization} model. 
Notice that the person identities do not overlap in training and testing split.
This shows that our large-scale dataset can help learn discriminative ReID features. 
Also, the fine-tuned model (\textbf{Adaptation} model) is superior to the model trained from scratch (\textbf{Supervised} model). 
Meanwhile, the performance of all models varies across different environments. 
All models perform poorly in \textit{Retail} environment due to its cluttered background. 
In general, the experiment shows that Re-ID is very challenging in cluttered and crowded environments in multi-camera settings. 
Our large-scale dataset can help learn a more discriminative Re-ID feature that is adapted to a given environment. However, the performance is still far from satisfactory in a challenging environment. We believe that more identities are needed to learn more discriminative Re-ID features in these challenging environments.


\begin{table}[htbp]
	\scriptsize
	\centering
	\begin{tabular}{c|cc|cc|cc}
		\toprule
		\multirow{2}[0]{*}{Env} & \multicolumn{2}{c|}{Generalization} & \multicolumn{2}{c|}{Adaptation} & \multicolumn{2}{c}{Supervised} \\
		\cmidrule{2-7}
		& mAP   & R-1 & mAP   & R-1 & mAP   & R-1\\
		\midrule
		Cafe & 48.82 & 77.78 & 63.61 & 88.01 & 59.55 & 87.80 \\
		Industry & 39.42 & 65.84 & 51.39 & 79.15 & 44.77 & 76.26 \\
		Lobby & 46.08 & 72.63 & 60.36 & 87.43 & 51.63 & 82.79 \\
		Office & 42.89 & 73.47 & 58.72 & 80.64 & 51.20 & 76.79\\
		Retail & 28.46 & 49.33 & 33.25 & 58.29 & 31.64 & 57.43 \\
		\bottomrule
	\end{tabular}%
	\caption{Person re-identification (ReID) performance of each environment on the testing split. We report the performance of three different training settings. } 
	\label{tab:reid}%
	\vspace{-15pt}
\end{table}%

\section{Conclusion}
\label{sec:conclusion}
In deep learning, high-quality labeled data is fundamental to achieving desired performance in many AI tasks.
This is particularly the case for multi-camera multiple people tracking, 
where trackers' performance is profoundly impacted by environment settings, \eg, camera angles, backgrounds, and lighting conditions.
In this work, We manage to build the largest multi-camera multiple people tracking dataset 
with the help of an auto-annotation system, which employs various calibrated depth sensors and RGB sensors to construct a robust 3D tracker and generates reliable multi-camera tracking ground truth.
Our dataset offers high-quality, dense annotations for all the captured frames. 
We study the performance of two real-time trackers and one robust ReID model on the proposed dataset.
The results suggest that a large-scale dataset allows the tracking systems and the ReID models to perform better. 
We believe these findings will benefit the design of real-world tracking systems.
For example, for a large chain retailer, where the interior designs are similar across different stores, 
we can use the auto-annotation system to create a large-scale multi-camera tracking dataset and train an adapted tracker using this dataset to achieve high accuracy for these environments.
On the other hand, the experiments also show the challenges of designing a highly accurate multi-camera tracking system 
in a cluttered and crowded environment, and the baseline methods are far from meeting the accuracy requirements of the demanding applications.
We hope our dataset can encourage more research efforts to be invested in this topic.

{\small
\bibliographystyle{ieee_fullname}
\bibliography{egbib}

\begin{thebibliography}{10}\itemsep=-1pt

\bibitem{amazongo}
Amazon go.
\newblock http://amazongo.com, 2017.

\bibitem{DBLP:journals/corr/Alameda-PinedaS15}
Xavier Alameda{-}Pineda, Jacopo Staiano, Ramanathan Subramanian, Ligia~Maria
  Batrinca, Elisa Ricci, Bruno Lepri, Oswald Lanz, and Nicu Sebe.
\newblock {SALSA:} {A} novel dataset for multimodal group behavior analysis.
\newblock {\em CoRR}, abs/1506.06882, 2015.

\bibitem{andriluka2008people}
Mykhaylo Andriluka, Stefan Roth, and Bernt Schiele.
\newblock People-tracking-by-detection and people-detection-by-tracking.
\newblock In {\em 2008 IEEE Conference on computer vision and pattern
  recognition}, pages 1--8. IEEE, 2008.

\bibitem{baque2017deep}
Pierre Baqu{\'e}, Fran{\c{c}}ois Fleuret, and Pascal Fua.
\newblock Deep occlusion reasoning for multi-camera multi-target detection.
\newblock In {\em Proceedings of the IEEE International Conference on Computer
  Vision}, pages 271--279, 2017.

\bibitem{berclaz2011multiple}
Jerome Berclaz, Francois Fleuret, Engin Turetken, and Pascal Fua.
\newblock Multiple object tracking using k-shortest paths optimization.
\newblock {\em IEEE transactions on pattern analysis and machine intelligence},
  33(9):1806--1819, 2011.

\bibitem{berman2019multigrain}
Maxim Berman, Herv{\'e} J{\'e}gou, Andrea Vedaldi, Iasonas Kokkinos, and
  Matthijs Douze.
\newblock Multigrain: a unified image embedding for classes and instances.
\newblock {\em arXiv preprint arXiv:1902.05509}, 2019.

\bibitem{bernardin2008evaluating}
Keni Bernardin and Rainer Stiefelhagen.
\newblock Evaluating multiple object tracking performance: the clear mot
  metrics.
\newblock {\em EURASIP Journal on Image and Video Processing}, 2008:1--10,
  2008.

\bibitem{chavdarova2018wildtrack}
Tatjana Chavdarova, Pierre Baqu{\'e}, St{\'e}phane Bouquet, Andrii Maksai, Cijo
  Jose, Timur Bagautdinov, Louis Lettry, Pascal Fua, Luc Van~Gool, and
  Fran{\c{c}}ois Fleuret.
\newblock Wildtrack: A multi-camera hd dataset for dense unscripted pedestrian
  detection.
\newblock In {\em Proceedings of the IEEE Conference on Computer Vision and
  Pattern Recognition}, pages 5030--5039, 2018.

\bibitem{chavdarova2017deep}
Tatjana Chavdarova and Fran{\c{c}}ois Fleuret.
\newblock Deep multi-camera people detection.
\newblock In {\em 2017 16th IEEE International Conference on Machine Learning
  and Applications (ICMLA)}, pages 848--853. IEEE, 2017.

\bibitem{chen2020multi}
He Chen, Pengfei Guo, Pengfei Li, Gim~Hee Lee, and Gregory Chirikjian.
\newblock Multi-person 3d pose estimation in crowded scenes based on multi-view
  geometry.
\newblock In {\em European Conference on Computer Vision}, pages 541--557.
  Springer, 2020.

\bibitem{chen2016equalized}
Weihua Chen, Lijun Cao, Xiaotang Chen, and Kaiqi Huang.
\newblock An equalized global graph model-based approach for multicamera object
  tracking.
\newblock {\em IEEE Transactions on Circuits and Systems for Video Technology},
  27(11):2367--2381, 2016.

\bibitem{dehghan2015gmmcp}
Afshin Dehghan, Shayan Modiri~Assari, and Mubarak Shah.
\newblock Gmmcp tracker: Globally optimal generalized maximum multi clique
  problem for multiple object tracking.
\newblock In {\em Proceedings of the IEEE conference on computer vision and
  pattern recognition}, pages 4091--4099, 2015.

\bibitem{eshel2008homography}
Ran Eshel and Yael Moses.
\newblock Homography based multiple camera detection and tracking of people in
  a dense crowd.
\newblock In {\em 2008 IEEE Conference on Computer Vision and Pattern
  Recognition}, pages 1--8. IEEE, 2008.

\bibitem{ferryman2009pets2009}
James Ferryman and Ali Shahrokni.
\newblock Pets2009: Dataset and challenge.
\newblock In {\em 2009 Twelfth IEEE international workshop on performance
  evaluation of tracking and surveillance}, pages 1--6. IEEE, 2009.

\bibitem{fleuret2007multicamera}
Francois Fleuret, Jerome Berclaz, Richard Lengagne, and Pascal Fua.
\newblock Multicamera people tracking with a probabilistic occupancy map.
\newblock {\em IEEE transactions on pattern analysis and machine intelligence},
  30(2):267--282, 2007.

\bibitem{he2020fastreid}
Lingxiao He, Xingyu Liao, Wu Liu, Xinchen Liu, Peng Cheng, and Tao Mei.
\newblock Fastreid: A pytorch toolbox for general instance re-identification.
\newblock {\em arXiv preprint arXiv:2006.02631}, 2020.

\bibitem{he2020multi}
Yuhang He, Xing Wei, Xiaopeng Hong, Weiwei Shi, and Yihong Gong.
\newblock Multi-target multi-camera tracking by tracklet-to-target assignment.
\newblock {\em IEEE Transactions on Image Processing}, 29:5191--5205, 2020.

\bibitem{hofmann2013hypergraphs}
Martin Hofmann, Daniel Wolf, and Gerhard Rigoll.
\newblock Hypergraphs for joint multi-view reconstruction and multi-object
  tracking.
\newblock In {\em Proceedings of the IEEE Conference on Computer Vision and
  Pattern Recognition}, pages 3650--3657, 2013.

\bibitem{hou2021multiview}
Yunzhong Hou and Liang Zheng.
\newblock Multiview detection with shadow transformer (and view-coherent data
  augmentation).
\newblock In {\em Proceedings of the 29th ACM International Conference on
  Multimedia}, pages 1673--1682, 2021.

\bibitem{hou2020multiview}
Yunzhong Hou, Liang Zheng, and Stephen Gould.
\newblock Multiview detection with feature perspective transformation.
\newblock In {\em European Conference on Computer Vision}, pages 1--18.
  Springer, 2020.

\bibitem{hou2019locality}
Yunzhong Hou, Liang Zheng, Zhongdao Wang, and Shengjin Wang.
\newblock Locality aware appearance metric for multi-target multi-camera
  tracking.
\newblock {\em arXiv preprint arXiv:1911.12037}, 2019.

\bibitem{hsu2020traffic}
Hung-Min Hsu, Yizhou Wang, and Jenq-Neng Hwang.
\newblock Traffic-aware multi-camera tracking of vehicles based on reid and
  camera link model.
\newblock In {\em Proceedings of the 28th ACM International Conference on
  Multimedia}, pages 964--972, 2020.

\bibitem{jiang2018online}
Na Jiang, SiChen Bai, Yue Xu, Chang Xing, Zhong Zhou, and Wei Wu.
\newblock Online inter-camera trajectory association exploiting person
  re-identification and camera topology.
\newblock In {\em Proceedings of the 26th ACM international conference on
  Multimedia}, pages 1457--1465, 2018.

\bibitem{glenn_jocher_2021_5563715}
Glenn Jocher, Alex Stoken, Ayush Chaurasia, Jirka Borovec, NanoCode012, TaoXie,
  Yonghye Kwon, Kalen Michael, Liu Changyu, Jiacong Fang, Abhiram V, Laughing,
  tkianai, yxNONG, Piotr Skalski, Adam Hogan, Jebastin Nadar, imyhxy, Lorenzo
  Mammana, AlexWang1900, Cristi Fati, Diego Montes, Jan Hajek, Laurentiu
  Diaconu, Mai~Thanh Minh, Marc, albinxavi, fatih, oleg, and wanghaoyang0106.
\newblock {ultralytics/yolov5: v6.0 - YOLOv5n 'Nano' models, Roboflow
  integration, TensorFlow export, OpenCV DNN support}, Oct. 2021.

\bibitem{kuo2010inter}
Cheng-Hao Kuo, Chang Huang, and Ram Nevatia.
\newblock Inter-camera association of multi-target tracks by on-line learned
  appearance affinity models.
\newblock In {\em European Conference on Computer Vision}, pages 383--396.
  Springer, 2010.

\bibitem{5979949}
Rainer Kümmerle, Giorgio Grisetti, Hauke Strasdat, Kurt Konolige, and Wolfram
  Burgard.
\newblock G2o: A general framework for graph optimization.
\newblock In {\em 2011 IEEE International Conference on Robotics and
  Automation}, pages 3607--3613, 2011.

\bibitem{law2018cornernet}
Hei Law and Jia Deng.
\newblock Cornernet: Detecting objects as paired keypoints.
\newblock In {\em Proceedings of the European conference on computer vision
  (ECCV)}, pages 734--750, 2018.

\bibitem{lima2021generalizable}
Joao~Paulo Lima, Rafael Roberto, Lucas Figueiredo, Francisco Simoes, and
  Veronica Teichrieb.
\newblock Generalizable multi-camera 3d pedestrian detection.
\newblock In {\em Proceedings of the IEEE/CVF Conference on Computer Vision and
  Pattern Recognition}, pages 1232--1240, 2021.

\bibitem{lin2014microsoft}
Tsung-Yi Lin, Michael Maire, Serge Belongie, James Hays, Pietro Perona, Deva
  Ramanan, Piotr Doll{\'a}r, and C~Lawrence Zitnick.
\newblock Microsoft coco: Common objects in context.
\newblock In {\em European conference on computer vision}, pages 740--755.
  Springer, 2014.

\bibitem{luo2019bag}
Hao Luo, Youzhi Gu, Xingyu Liao, Shenqi Lai, and Wei Jiang.
\newblock Bag of tricks and a strong baseline for deep person
  re-identification.
\newblock In {\em Proceedings of the IEEE/CVF Conference on Computer Vision and
  Pattern Recognition Workshops}, pages 0--0, 2019.

\bibitem{luo2020multiple}
Wenhan Luo, Junliang Xing, Anton Milan, Xiaoqin Zhang, Wei Liu, and Tae-Kyun
  Kim.
\newblock Multiple object tracking: A literature review.
\newblock {\em Artificial Intelligence}, page 103448, 2020.

\bibitem{ristani2016performance}
Ergys Ristani, Francesco Solera, Roger Zou, Rita Cucchiara, and Carlo Tomasi.
\newblock Performance measures and a data set for multi-target, multi-camera
  tracking.
\newblock In {\em European conference on computer vision}, pages 17--35.
  Springer, 2016.

\bibitem{ristani2018features}
Ergys Ristani and Carlo Tomasi.
\newblock Features for multi-target multi-camera tracking and
  re-identification.
\newblock In {\em Proceedings of the IEEE conference on computer vision and
  pattern recognition}, pages 6036--6046, 2018.

\bibitem{song2021stacked}
Liangchen Song, Jialian Wu, Ming Yang, Qian Zhang, Yuan Li, and Junsong Yuan.
\newblock Stacked homography transformations for multi-view pedestrian
  detection.
\newblock In {\em Proceedings of the IEEE/CVF International Conference on
  Computer Vision}, pages 6049--6057, 2021.

\bibitem{tesfaye2017multi}
Yonatan~Tariku Tesfaye, Eyasu Zemene, Andrea Prati, Marcello Pelillo, and
  Mubarak Shah.
\newblock Multi-target tracking in multiple non-overlapping cameras using
  constrained dominant sets.
\newblock {\em arXiv preprint arXiv:1706.06196}, 2017.

\bibitem{tu2020voxelpose}
Hanyue Tu, Chunyu Wang, and Wenjun Zeng.
\newblock Voxelpose: Towards multi-camera 3d human pose estimation in wild
  environment.
\newblock In {\em Proceedings of the European Conference on Computer Vision
  (ECCV)}, pages 197--212. Springer, 2020.

\bibitem{vo2020self}
Minh~Phuoc Vo, Ersin Yumer, Kalyan Sunkavalli, Sunil Hadap, Yaser~A Sheikh, and
  Srinivasa~G Narasimhan.
\newblock Self-supervised multi-view person association and its applications.
\newblock {\em IEEE transactions on pattern analysis and machine intelligence},
  2020.

\bibitem{wan2013distributed}
Jiuqing Wan and Liu Li.
\newblock Distributed optimization for global data association in
  non-overlapping camera networks.
\newblock In {\em 2013 Seventh International Conference on Distributed Smart
  Cameras (ICDSC)}, pages 1--7. IEEE, 2013.

\bibitem{wang2018non}
Xiaolong Wang, Ross Girshick, Abhinav Gupta, and Kaiming He.
\newblock Non-local neural networks.
\newblock In {\em Proceedings of the IEEE conference on computer vision and
  pattern recognition}, pages 7794--7803, 2018.

\bibitem{wei2018person}
Longhui Wei, Shiliang Zhang, Wen Gao, and Qi Tian.
\newblock Person transfer gan to bridge domain gap for person
  re-identification.
\newblock In {\em Proceedings of the IEEE conference on computer vision and
  pattern recognition}, pages 79--88, 2018.

\bibitem{wen2017multi}
Longyin Wen, Zhen Lei, Ming-Ching Chang, Honggang Qi, and Siwei Lyu.
\newblock Multi-camera multi-target tracking with space-time-view hyper-graph.
\newblock {\em International Journal of Computer Vision}, 122(2):313--333,
  2017.

\bibitem{xu2016multi}
Yuanlu Xu, Xiaobai Liu, Yang Liu, and Song-Chun Zhu.
\newblock Multi-view people tracking via hierarchical trajectory composition.
\newblock In {\em Proceedings of the IEEE Conference on Computer Vision and
  Pattern Recognition}, pages 4256--4265, 2016.

\bibitem{xu2019unified}
Yuhao Xu and Jiakui Wang.
\newblock A unified neural network for object detection, multiple object
  tracking and vehicle re-identification.
\newblock {\em arXiv preprint arXiv:1907.03465}, 2019.

\bibitem{you2019action4d}
Quanzeng You and Hao Jiang.
\newblock Action4d: Online action recognition in the crowd and clutter.
\newblock In {\em Proceedings of the IEEE/CVF Conference on Computer Vision and
  Pattern Recognition}, pages 11857--11866, 2019.

\bibitem{you2020real}
Quanzeng You and Hao Jiang.
\newblock Real-time 3d deep multi-camera tracking.
\newblock {\em arXiv preprint arXiv:2003.11753}, 2020.

\bibitem{zamir2012gmcp}
Amir~Roshan Zamir, Afshin Dehghan, and Mubarak Shah.
\newblock Gmcp-tracker: Global multi-object tracking using generalized minimum
  clique graphs.
\newblock In {\em European conference on computer vision}, pages 343--356.
  Springer, 2012.

\bibitem{zhang2015camera}
Shu Zhang, Elliot Staudt, Tim Faltemier, and Amit~K Roy-Chowdhury.
\newblock A camera network tracking (camnet) dataset and performance baseline.
\newblock In {\em 2015 IEEE Winter Conference on Applications of Computer
  Vision}, pages 365--372. IEEE, 2015.

\bibitem{zhang2021voxeltrack}
Yifu Zhang, Chunyu Wang, Xinggang Wang, Wenyu Liu, and Wenjun Zeng.
\newblock Voxeltrack: Multi-person 3d human pose estimation and tracking in the
  wild.
\newblock {\em arXiv preprint arXiv:2108.02452}, 2021.

\bibitem{zhang2021fairmot}
Yifu Zhang, Chunyu Wang, Xinggang Wang, Wenjun Zeng, and Wenyu Liu.
\newblock Fairmot: On the fairness of detection and re-identification in
  multiple object tracking.
\newblock {\em International Journal of Computer Vision}, pages 1--19, 2021.

\bibitem{zheng2016person}
Liang Zheng, Yi Yang, and Alexander~G Hauptmann.
\newblock Person re-identification: Past, present and future.
\newblock {\em arXiv preprint arXiv:1610.02984}, 2016.

\end{thebibliography}
}

\end{document}